%% file: manuscript_with_title.tex
\documentclass[a4paper,fleqn]{cas-dc}
\usepackage[numbers]{natbib}
\usepackage{threeparttable}
\usepackage{adjustbox}
\usepackage{tikz}
\usepackage{docmute}

\def\tsc#1{\csdef{#1}{\textsc{\lowercase{#1}}\xspace}}
\tsc{WGM}
\tsc{QE}
\tsc{EP}
\tsc{PMS}
\tsc{BEC}
\tsc{DE}
\newcommand{\greenup}{\textcolor{green}{$\uparrow$}}
\newcommand{\reddown}{\textcolor{red}{$\downarrow$}}

\begin{document}

\newgeometry{margin=2.5cm}
\begingroup
\fontfamily{lmr}\fontsize{11}{13.6}\selectfont
\setlength{\parindent}{0pt}
\setlength{\parskip}{0.65em}
\thispagestyle{empty}

\input{title_page.tex}

\par\clearpage
\endgroup
\restoregeometry

\input{manuscript_anonymized.tex}

\end{document}

%% file: title_page.tex
\begin{center}
{\LARGE\bfseries Towards Scalable Diagnostic-Driven Whole-Slide Image
Retrieval for Pathology Archives\par}

\vspace{1.5em}
{\large Jingsong Liu\textsuperscript{1,3,6,*,$\dagger$},
Youssef Wally\textsuperscript{2,$\dagger$},
Han Li\textsuperscript{1,3,4},
Weiwei Zhou \textsuperscript{5},
Jing Dai\textsuperscript{5},
Elisabeth Wetzer\textsuperscript{2},
Peter J. Schüffler\textsuperscript{1,3,6}\par}
\end{center}

\textsuperscript{1}Institute of Pathology, Technical University of Munich, TUM School of Medicine
and Health, Munich, Germany

\textsuperscript{2}Department of Physics and Technology, UiT The Arctic University of Norway, Tromsø, Norway

\textsuperscript{3}Munich Center for Machine Learning (MCML), Munich, Germany

\textsuperscript{4}Computer Aided Medical Procedures (CAMP), Technical University of Munich, Munich, Germany

\textsuperscript{5}School of Biomedical Engineering, Faculty of Medicine, Dalian University of Technology, Dalian, China

\textsuperscript{6}Munich Data Science Institute (MDSI), Munich, Germany

\textsuperscript{$\dagger$}These authors contributed equally to this work.

\textsuperscript{*}\textbf{Corresponding author:} Jingsong Liu\\
\textbf{Email:} jingsong.liu@tum.de\\

\section*{Acknowledgements}
\input{components/acknowledgment}

\section*{CRediT authorship contribution statement}
\textbf{Jingsong Liu:} Conceptualization, Methodology, Writing.
\textbf{Youssef Wally:} Methodology, Validation, Original Draft.
\textbf{Han Li:} Writing.
\textbf{Weiwei Zhou:} Writing.
\textbf{Jing Dai:} Writing.
\textbf{Elisabeth Wetzer:} Writing and Supervision.
\textbf{Peter J. Schüffler:} Conceptualization, Supervision, Funding acquisition, Review and Editing.

%% file: components/acknowledgment.tex
This work was supported by the German Federal Ministry of Education and Research (BMBF)-funded SATURN3 project (01KD2206C) and the Innovative Medicines Initiative (IMI) BIGPICTURE project (IMI945358), the Research Council of Norway, Integreat – Norwegian Centre for knowledge-driven machine learning (project number 332645).
This work was supported through the Bavarian AI Foundation Model Initiative (AI-Bay.eu), funded by the Bavarian State Ministry of Science and the Arts (StMWK).

%% file: manuscript_anonymized.tex
\input{main.tex}

%% file: main.tex
\let\WriteBookmarks\relax
\def\floatpagepagefraction{1}
\def\textpagefraction{.001}
\shorttitle{Towards Scalable Diagnostic-Driven Whole-Slide Image Retrieval}
\shortauthors{Anonymous manuscript}

\title [mode = title]{Towards Scalable Diagnostic-Driven Whole-Slide Image Retrieval for Pathology Archives}      



\begin{abstract}
The growing adoption of digital pathology has rapidly expanded collections of hematoxylin-and-eosin whole-slide images, yet efficient navigation through these archives remains challenging, as metadata is often inconsistent or unavailable. Content-based image retrieval addresses this challenge by identifying similar cases directly from visual representations instead of manual annotations or metadata. Existing retrieval systems have demonstrated promising performance but largely rely on handcrafted indexing pipelines and predefined sampling strategies, constraining scalability and generalization.
Here, we present CLEAR-WSI (Constant-Length Embedding for Automatic Retrieval of Whole-Slide Images) as a scalable, diagnostic-driven slide-retrieval framework, validated on both public and clinical data.
Using a fixed-length slide embedding, CLEAR-WSI refines retrieved candidates with either a predicted diagnostic labdddel or the ground-truth label, depending on the retrieval setting.
Newly processed cases are subsequently added to the expanding memory bank, allowing the retrieval archive to grow over time.
On a comprehensive benchmark consisting of two public and one private clinical datasets representing four retrieval tasks, CLEAR-WSI improves $\mathrm{Precision}@5$ over the strongest evaluated baseline, with gains on CAMELYON16 (3.10), BRACS-1 (2.75), BRACS-2 (1.95), and X-HER2 (3.75). Additional ablation and latency studies further examine the contribution of each component and highlight its potential for scalable deployment.
\end{abstract}

\begin{graphicalabstract}
\includegraphics[width=\textwidth,keepaspectratio]{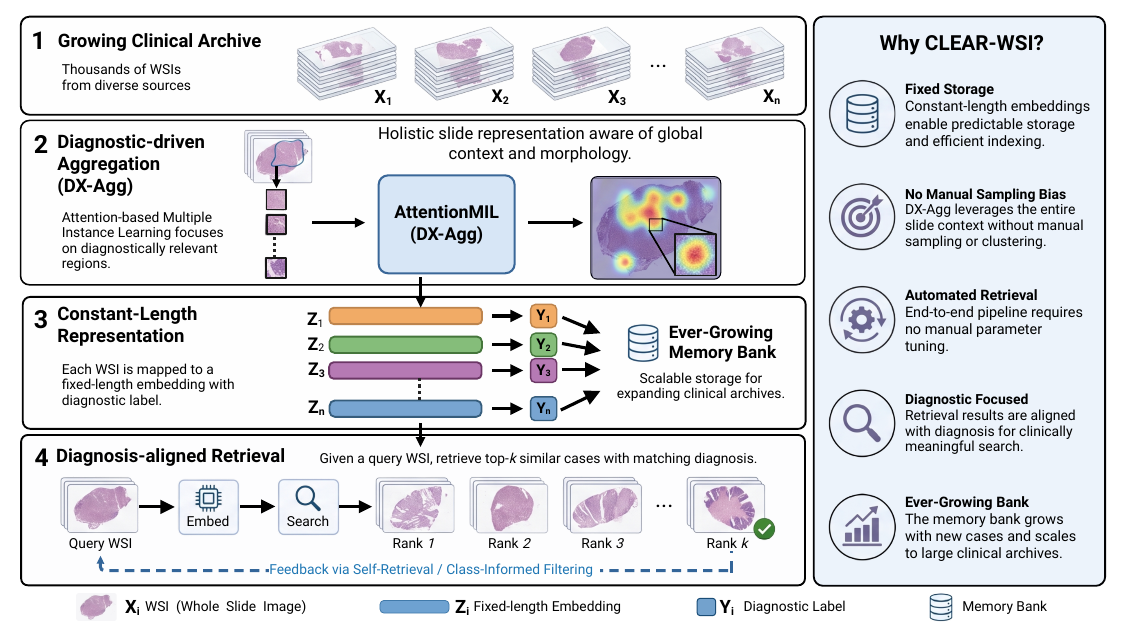}
\end{graphicalabstract}

\begin{highlights} 
\item CLEAR-WSI combines Diagnostic-aligned Aggregation, constant-length embeddings with a growing memory bank, and both Class-Informed and Self-Retrieval configurations for flexible retrieval with or without available diagnoses. 
\item CLEAR-WSI achieves state-of-the-art retrieval performance on a comprehensive benchmark across three public and one private clinical tasks of varying difficulty.
\item Additional ablation and latency studies further examine the contribution of each component and demonstrate the framework's potential for scalable deployment. 
\end{highlights}

\begin{keywords}
Image Retrieval \sep Content-Based Image Retrieval \sep Whole Slide Images \sep Digital Pathology \sep Foundation Models
\end{keywords}

\maketitle

\section{Introduction}
\label{sec:introduction}
\input{components/introduction}

\section{Related Work}
\label{sec:related_work}
\input{components/related_works}

\section{Methods}
\label{sec:methods}
\input{components/methods}

\section{Experiments}
\label{sec:experiments}
\input{components/experiments}

\section{Ablation and Latency Study}
\label{sec:ablation}
\input{components/ablation_study}

\section{Conclusion and Discussion}
\label{sec:discussion}
\input{components/discussion}

\clearpage
\appendix
\setcounter{table}{0}
\renewcommand{\thetable}{S\arabic{table}}
\section{Complete Numerical Retrieval Results}
\label{app:complete_results}
\input{components/appendix}
\clearpage

\bibliographystyle{cas-model2-names}
\bibliography{cas-refs}





%% file: components/introduction.tex
Digital pathology archives are currently growing, thus complicating the
retrieval of relevant slides for a particular query slide.
Metadata-based searches are often hindered by inconsistent or incomplete
databases.
Alternatively, content-based image retrieval (CBIR) enables searching for
whole-slide images (WSIs) directly from a query image, but requires a useful
representation and similarity measure between slides.
For this work, we assume that the main objective is to find diagnostically
related cases in the archive.
Such retrieval can support case-based review \cite{kalra2020yottixel},
education \cite{chen2022fast}, and similarity-driven exploration
\cite{Zheng2003}.
However, WSIs are gigapixel-sized images that contain a variable number of
heterogeneous tissue regions, making conventional methods inefficient and
challenging to apply \cite{li2023segment, xu2026pathpilot, xu2026mmnavagent}.
A retrieval system must therefore represent diagnostically relevant
morphology in a form that can be efficiently stored and searched.

Existing CBIR approaches address the gigapixel scale by representing each slide
with a selected set of local observations.
Earlier systems used handcrafted color and texture descriptors
\cite{Zheng2003}; recent WSI systems instead use learned patch features within
mosaic-based retrieval \cite{kalra2020yottixel}, slide-level hashing
\cite{li2023high}, discrete-code indexing \cite{chen2022fast}, or local patch
matching \cite{wang2023retccl}.
While these approaches improve retrieval efficiency, they also involve
trade-offs in how a slide is represented.
The number of stored patches may depend on tissue content and sampling
decisions, whereas stronger compression can remove diagnostically relevant
details.
In addition, visually similar slides are not necessarily diagnostically
consistent.

For a scalable, searchable WSI archive, we argue that (i) the storage required for a newly indexed WSI should not depend on how many patches it contains, (ii) a WSI's representation must preserve diagnostic evidence without relying on manually configured clustering or mosaic construction, and (iii) the search procedure must accommodate queries both with and without an available diagnosis.
Motivated by these considerations, we recently conceptualized the diagnosis-aligned WSI retrieval framework \textit{CLEAR-WSI} (Constant-Length Embedding for Automatic Retrieval of Whole-Slide Images) for growing clinical archives \cite{WalYou_CLEARWSI_MICCAISAT2026}.

\begin{figure*}[t]
    \centering
    \includegraphics[width=1.0\textwidth]{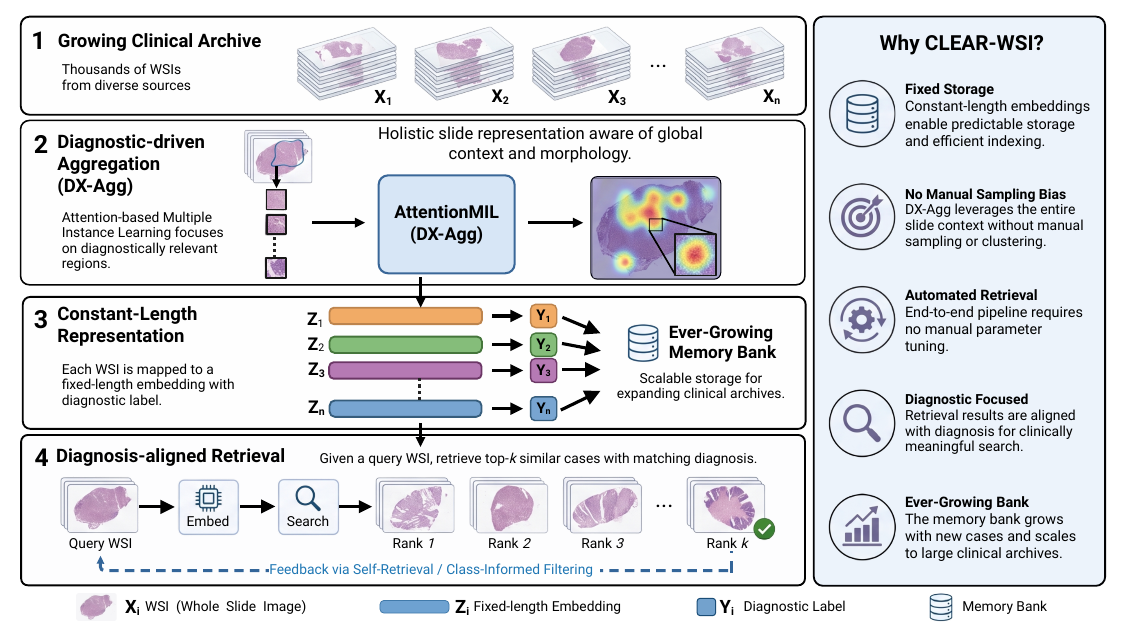}
    \caption{\textbf{Overview of the \textit{CLEAR-WSI} framework for scalable,
    diagnosis-aligned retrieval from growing clinical archives.}
    Left: (1) A growing clinical archive contains WSIs from heterogeneous sources.
    (2) Each WSI is processed by Diagnostic-driven Aggregation (DX-Agg), implemented
    with Attention-based MIL, to identify diagnostically relevant evidence and capture
    global slide morphology.
    (3) DX-Agg maps each variable-size patch set to a fixed-length slide embedding $Z_i$
    and diagnostic prediction $Y_i$, which are stored in an ever-growing memory bank,
    providing constant per-slide storage and predictable index expansion.
    (4) Given a query WSI, \textit{CLEAR-WSI} retrieves the top-$k$ similar cases and
    refines the ranked list using either Self-Retrieval (SR) filtering, when the query
    diagnosis is unknown, or Class-Informed (CI) filtering, when it is available.
    Right: Key features of \textit{CLEAR-WSI}, including fixed-length storage,
    automated diagnosis-focused retrieval, and a scalable, ever-growing memory bank.}
    \label{fig:figure1}
\end{figure*}

In this work, we review this concept and extend it with a grid-search strategy for improved hyperparameter selection, dedicated ablation and latency studies, and a clinical evaluation study.
Following its initial overall structure, CLEAR-WSI can be formalized as shown in Figure~\ref{fig:figure1}. In a nutshell, it
uses a diagnosis-aware, fixed-length
slide representation that provides a constant per-slide storage footprint and
predictable memory-bank growth, and integrates automated Self-Retrieval (SR)
and label-available Class-Informed (CI) modes to support diagnosis-aligned
search both with and without an available query diagnosis.
We show on two public datasets (CAMELYON16, BRACS-1/2) and a private clinical
X-HER2 cohort that \textit{CLEAR-WSI} consistently outperforms representative
state-of-the-art retrieval methods, making it a powerful tool for archival
slide retrieval. 
 \textit{CLEAR-WSI} is openly available at \href{https://anonymous.4open.science/r/CLEAR-WSI-Plus-4012/README.md}{Anonymized Git Repo}.

%% file: components/related_works.tex
Existing approaches can be broadly organized according to how they represent a complete slide and perform retrieval: mosaic or patch-set matching, discrete-code indexing, slide-level hashing, and retrieval using foundation-model representations.

\subsection{Mosaic or Patch-Set Retrieval}

Early CBIR systems represented histopathology images using handcrafted color, texture, and local structural descriptors \cite{Zheng2003}. Although these methods enabled example-based search, they were primarily developed for small images or manually selected regions and therefore did not directly address the gigapixel scale of WSIs. Subsequent WSI retrieval systems reduce this complexity by selecting a representative set of patches, often referred to as a \emph{mosaic}, and deriving slide similarity from patch-level comparisons.
Yottixel is a representative mosaic-based search engine that clusters tissue regions by color, spatially samples representative patches, and encodes them for efficient matching \cite{kalra2020yottixel}. Its mosaic substantially reduces the number of patches that must be stored and compared, but the resulting slide representation remains dependent on clustering and sampling choices. Retrieval with Clustering-Guided Contrastive Learning (RetCCL) improves patch representations through clustering-guided contrastive learning and selects distinctive query patches using feature- and spatial-clustering \cite{wang2023retccl}. It performs patch-by-patch retrieval and subsequently aggregates the local matches to rank their source WSIs. These approaches retain local evidence and can highlight corresponding regions, which supports interpretation of the retrieved results. 

However, a complete WSI is represented by a collection of selected patch descriptors rather than a single global vector. Consequently, storage, matching complexity, and retrieval behavior remain coupled to the construction and size of the representative patch set, while potentially important regions may be omitted during sampling. This variable-size slide representation leads to an unpredictable storage cost per case and complicates downstream pipelines and capacity planning during deployment in continuously expanding clinical archives \cite{hanna2022integrating}.

\subsection{Discrete-Code and Indexed Patch-Set Retrieval}

Discrete-code methods improve retrieval efficiency by mapping continuous image
features to compact integer or binary representations that can be searched
using specialized indices.
Earlier histopathology retrieval studies used supervised kernel hashing
\cite{Zhang2015} and deep ranking objectives \cite{NDCGOG} to reduce storage
and accelerate similarity search.
Nevertheless, extending these techniques from individual images or patches to
heterogeneous WSIs requires an additional mechanism for organizing and
comparing multiple local representations.

Self-Supervised Image Search for Histology (SISH) addresses this problem by combining mosaic construction with self-supervised discrete representation learning \cite{chen2022fast}. A vector-quantized variational autoencoder (VQ-VAE) maps each representative patch to a discrete latent code, and the resulting patch indices are organized in a van Emde Boas tree for efficient candidate search, followed by uncertainty-based ranking. SISH therefore achieves fast indexed search and retains patch-level interpretability.

Importantly, its fixed-size patch codes should not be confused with a
fixed-length slide embedding: each WSI is still indexed as a set of mosaic
patches and their associated codes.
Accordingly, SISH compresses and indexes local entries efficiently but does
not provide a uniform one-vector-per-slide representation.

\subsection{Slide-Level Hashing Retrieval}

Slide-level hashing methods seek to incorporate global WSI information while preserving the storage and search advantages of compact binary codes. High-Order Correlation-Guided Slide-Level Histology Retrieval with Self-Supervised Hashing (HSHR) learns an attention-based hash encoder that captures high-order relationships among patch features and generates compact hash representations for slide cluster centers \cite{li2023high}. Compared with direct patch matching, this formulation introduces slide-level context and uses learned attention to emphasize informative tissue patterns.

However, quantization into binary codes may discard fine-grained morphological information, and the cluster-center representation retains multiple codes rather than mapping every slide to one continuous embedding. This design requires multi-component matching instead of a uniform one-vector-per-slide index. These trade-offs motivate representations that retain compact slide-level indexing without imposing discrete quantization or patch-set comparison.

\subsection{Foundation Models for Histopathology Retrieval}

Recent pathology foundation models (FMs) learn transferable, domain-specific
representations from large WSI collections
\cite{vorontsov2024foundation,wang2022transformer,filiot2024phikon,
liu2026cellularscaleinterpretabilitypathologyfoundation}.
Most operate at the patch level; for example, UNI was pretrained on over
100 million patches from more than 100,000 diagnostic WSIs
\cite{chen2024uni}.
Although these embeddings improve local retrieval, patch-level foundation
models do not by themselves define a slide-level retrieval representation;
an additional aggregation or patch-matching mechanism is still required.
A smaller but growing class of whole-slide foundation models explicitly aggregates local features into general-purpose slide representations. Prov-GigaPath provides a pretrained slide encoder, while the multimodal TITAN model has demonstrated rare-cancer and cross-modal retrieval capabilities \cite{xu2024gigapath,ding2025multimodal}. However, these models are designed as broadly transferable representation learners rather than retrieval systems explicitly optimized for diagnosis-aligned indexing.

In contrast, \textit{CLEAR-WSI} couples pathology FM features with Diagnostic-driven Aggregation (DX-Agg) to map every WSI to a fixed-length slide embedding, irrespective of its tissue area or number of extracted patches, and applies SR filtering to improve diagnostic consistency. This provides a fixed storage cost for each newly indexed slide, predictable linear growth of the memory bank, and uniform vector-to-vector matching. These properties simplify storage and compute planning, while the latency analysis in this study evaluates retrieval efficiency at the tested memory-bank scale.

%% file: components/methods.tex
\noindent\textbf{System Formulation and State.}
The workflow in Fig.~\ref{fig:figure1} separates slide encoding from archive search. At any retrieval step $t$, the system maintains a memory bank
\begin{equation}
\mathcal{M}^{(t)}=\{(Z_i,\tilde{y}_i)\}_{i=1}^{N_t},
\end{equation}
where $t$ denotes the retrieval step, $N_t=|\mathcal{M}^{(t)}|$ is the number of indexed WSIs, $i$ indexes memory-bank entries, and $d$ is the embedding dimensionality. Here, $Z_i\in\mathbb{R}^{d}$ is a fixed-length WSI embedding and $\tilde{y}_i$ is its predicted diagnostic label. This representation gives every indexed slide the same vector storage cost. Processing a query comprises four operations: construction of its slide representation, similarity ranking, optional diagnosis-aware filtering, and insertion of the processed query into $\mathcal{M}^{(t)}$.

\noindent\textbf{From Tissue Patches to a Slide Representation.}
Because direct processing at native WSI resolution is computationally prohibitive \cite{liu2025hasd}, each slide $X_l$ is first tissue-segmented and tessellated using the standard WSI preprocessing workflow \cite{lu2021data}. The retained tissue patches form a variable-size bag
\begin{equation}
X_l=\{p_{lj}\}_{j=1}^{N_l}, \qquad p_{lj}\in\mathbb{R}^{h\times w\times c}.
\end{equation}
A patch encoder $f_\theta$, instantiated by a pathology foundation model such as UNI \cite{chen2024uni}, maps every patch to $h_{lj}=f_\theta(p_{lj})\in\mathbb{R}^{d}$. Thus, differences in tissue area affect the number $N_l$ of local features but not their dimensionality.

DX-Agg converts this bag into a single slide vector. It is implemented by an AttentionMIL module \cite{ilse2018attention} trained using slide-level diagnostic supervision. For slide $l$, the module produces normalized non-negative weights $\{\beta_{lj}\}_{j=1}^{N_l}$, with $\sum_j\beta_{lj}=1$, and computes
\begin{equation}
Z_l=\sum_{j=1}^{N_l}\beta_{lj}h_{lj}.
\label{eq:dxagg}
\end{equation}
The weights allow regions that support the diagnostic objective to contribute more strongly, while Eq.~\ref{eq:dxagg} always returns one $d$-dimensional vector. A classification head $g_\phi$ applied to the same representation yields posterior probabilities and the predicted label
\begin{equation}
\tilde{y}_l=\arg\max_c\,[g_\phi(Z_l)]_c.
\end{equation}
The embedding is used for visual matching, whereas the prediction is reserved for label-consistency filtering and memory-bank storage.

\noindent\textbf{Memory-Bank Initialization and Similarity Search.}
Before test-time retrieval, the trained encoder and DX-Agg module are applied to the validation slides to initialize $\mathcal{M}^{(0)}$. Each entry stores the resulting embedding and predicted label; ground-truth labels are not stored as candidate labels in the retrieval bank. For a query slide $X_q$, the same encoding path produces $(Z_q,\tilde{y}_q)$. Its similarity to entry $i$ is cosine similarity,
\begin{equation}
s(q,i)=\frac{Z_q^\top Z_i}{\lVert Z_q\rVert_2\lVert Z_i\rVert_2},
\end{equation}
and the unfiltered ordering is
\begin{equation}
\pi(q)=\operatorname{argsort}_{i\in\{1,\ldots,N_t\}}s(q,i),
\end{equation}
with scores sorted in descending order.

\noindent\textbf{Two Diagnosis-added Filters.}
Filtering is applied after similarity ranking, so the underlying embedding space and similarity function remain identical in both modes. When the query diagnosis $y_q$ is available, CI retains candidates whose predicted bank label matches that diagnosis:
\begin{equation}
\pi_{\mathrm{CI}}(q)=\{i\in\pi(q)\mid \tilde{y}_i=y_q\}.
\end{equation}
CI is treated as a label-available oracle in our experiments rather than as the default automated configuration. When $y_q$ is unavailable, SR instead uses the query prediction:
\begin{equation}
\pi_{\mathrm{SR}}(q)=\{i\in\pi(q)\mid \tilde{y}_i=\tilde{y}_q\}.
\end{equation}
The returned neighborhood is formed by the first $k$ entries of the selected ordering. Test ground-truth labels are used only to score retrieval performance and, for CI, to define the explicit oracle query condition.

\noindent\textbf{Growing Memory-Bank.}
After the current retrieval is completed, the query becomes available to subsequent searches:
\begin{equation}
\mathcal{M}^{(t+1)}=\mathcal{M}^{(t)}\cup\{(Z_q,\tilde{y}_q)\}.
\end{equation}
This update increases the candidate pool by exactly one constant-length embedding and one predicted label per processed WSI. It does not update model parameters online. The growing-bank experiment therefore measures the effect of accumulating searchable cases, separately from representation learning and filter choice; a fixed-bank comparator keeps $\mathcal{M}^{(t)}=\mathcal{M}^{(0)}$ throughout evaluation.

%% file: components/experiments.tex
\definecolor{diffusionbg}{RGB}{230, 245, 255}
\newcommand{\bluebox}{\tikz\draw[blue,fill=diffusionbg] (0,0) rectangle (0.15,0.15);}

\subsection{Datasets}
Experiments were conducted on a comprehensive benchmark, consisting of two public and one private clinical whole-slide image datasets representing four retrieval tasks of different diagnostic complexity. \textbf{CAMELYON16} \cite{101093gigasciencegiy065} contains WSIs for lymph-node metastasis assessment and consists of two classes: \emph{normal} and \emph{tumor}. \textbf{BRACS} \cite{baac093} is a breast-cancer subtyping dataset containing seven diagnostic categories: Normal (N), Pathological Benign (PB), Usual Ductal Hyperplasia (UDH), Flat Epithelial Atypia (FEA), Atypical Ductal Hyperplasia (ADH), Ductal Carcinoma In Situ (DCIS), and Invasive Carcinoma (IC). Following the official benchmark protocol, retrieval is evaluated at two levels of diagnostic granularity. \textbf{BRACS-1} groups slides into three categories: Benign (N, PB), Atypical (UDH, FEA, ADH), and Malignant (DCIS, IC). \textbf{BRACS-2} uses the original seven-class subtype labels, representing a more fine-grained retrieval task. \textbf{X-HER2} is a private clinical breast-cancer cohort from X Hospital containing 77 WSIs. It is evaluated as a binary HER2-status retrieval task with two classes: \emph{HER2-negative} and \emph{HER2-positive}.

Predefined train, validation, and test splits were adopted for all experiments. The corresponding partitions contain 243, 27, and 129 WSIs for CAMELYON16; 393, 64, and 87 WSIs for BRACS; and 43, 18, and 16 WSIs for X-HER2, respectively.

\subsection{Baselines} 
\noindent\textbf{Yottixel-K} \cite{kalra2020yottixel} constructs a representative patch mosaic for each WSI, converts patch features into binary barcodes, and ranks slides using barcode-based Hamming distances. We reproduced Yottixel-K using KimiaNet with its default tissue threshold of 0.7. \textbf{SISH} \cite{chen2022fast} encodes mosaic patches using VQ-VAE-based semantic codes and DenseNet-121-based binary texture codes, which are indexed in a van Emde Boas tree for efficient retrieval. We reproduced SISH using the released VQ-VAE checkpoint and semantic codebook with the default Hamming-distance threshold of 128. \textbf{RetCCL} \cite{wang2023retccl} constructs representative patch mosaics using contrastively learned ResNet-50 features and produces slide-level retrieval results through cosine-similarity matching, entropy-based filtering, and voting. We reproduced RetCCL using the released checkpoint with nine feature clusters, a 20\% sampling ratio per cluster, and the default cosine-similarity threshold of 0.7.  
All baselines were reproduced using the publicly available implementations provided by the PathologySearchComparison benchmark.\footnote{\url{https://github.com/jacobluber/PathologySearchComparison}}

\subsection{CLEAR-WSI Setup}

\noindent\textbf{Encoder Selection.}
We compare four Vision Transformer (ViT)-based encoders: the general-domain
Data-efficient Image Transformer (DeiT) \cite{pmlr-v139-touvron21a} and
Momentum Contrast version 3 (MoCo v3) \cite{Chen2021b}, and the
pathology-pretrained Prov-GigaPath tile encoder \cite{xu2024gigapath} and
UNI \cite{chen2024uni}.
DeiT and MoCo v3 are pretrained on ImageNet-1K and are fine-tuned on
CAMELYON16 \cite{bejnordi2017diagnostic} and BRACS \cite{baac093}.
Prov-GigaPath was pretrained on approximately 1.3 billion tiles from more
than 170,000 WSIs, while UNI was pretrained on more than 100 million tissue
patches from over 100,000 diagnostic WSIs; both are kept frozen in our
experiments.
This comparison evaluates the effect of general-domain versus pathology-specific
patch representations on WSI retrieval.
Based on this comparison, UNI is selected as our encoder due to its superior
performance.

\noindent\textbf{Training.} 
An AttentionMIL model is trained as the Diagnostic-driven Aggregation (DX-Agg) module to map patch features to fixed-length slide embeddings and predict slide labels. The trained model is applied to the validation set to initialize the growing memory bank. For each validation slide, the bank stores its embedding and predicted label.

\noindent\textbf{Inference.} 
During inference, query slides are processed using either \textbf{Class-Informed (CI)} or \textbf{Self-Retrieval (SR)} filtering. CI uses the query ground-truth label and is treated as an oracle, label-available setting, whereas SR uses the label predicted by DX-Agg and requires no query annotation. The proposed system uses a growing memory bank: after retrieval, each query is added to the bank with its fixed-length slide embedding and predicted label. The fixed-bank configuration is evaluated only as a comparator in the ablation study.
 
\subsection{Evaluation Metrics}
Retrieval performance is quantified by evaluating how accurately the pipeline ranks relevant WSIs, where a retrieved slide is considered relevant if its label matches that of the query. Since clinically useful retrieval requires not only correct matches but also appropriate ranking of those matches, evaluation is performed using two complementary metrics: \textbf{(1) Precision:} We report $\mathrm{Precision}@k$ for $k \in \{1,3,5\}$, defined as the proportion of relevant slides among the top-$k$ retrieved results. \textbf{(2) Normalized Discounted Cumulative Gain (NDCG):} NDCG \cite{jarvelin2002cumulated} is widely adopted in recommender-system evaluation \cite{mao2021simplex,bellogin2011precision,xue2017deepmf,he2020lightgcn,rendle2009bpr,he2017neural}, but is less commonly used for WSI retrieval. Unlike Precision, NDCG accounts for the rank positions of relevant slides and rewards methods that place correct matches earlier. Following \cite{NDCGOG}, we therefore report both $\mathrm{Precision}@k$ and $\mathrm{NDCG}@k$.

\subsection{Retrieval Results}

\begin{figure*}[t]
    \centering
    \includegraphics[width=0.9\textwidth]{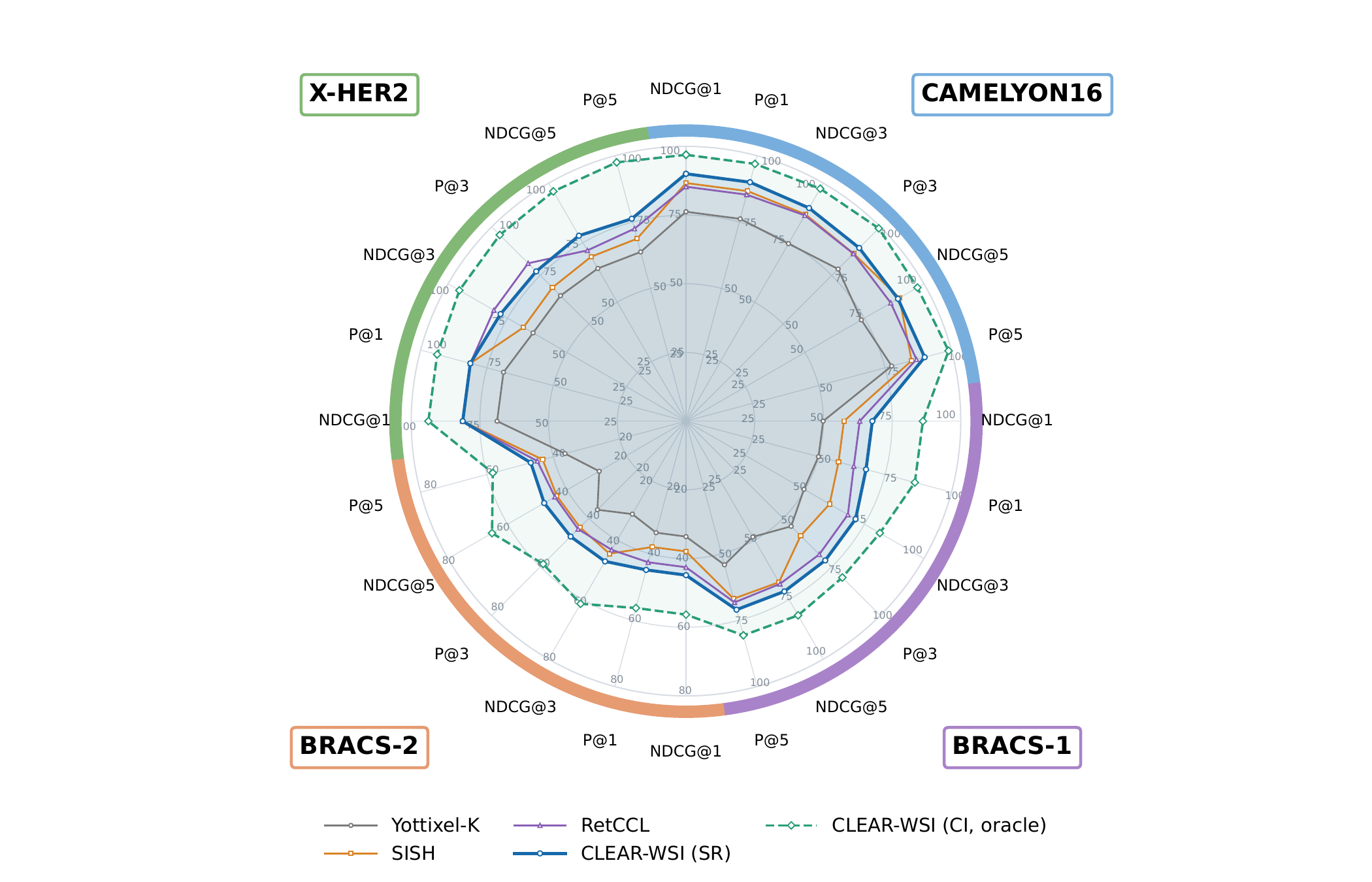}
    \caption{\textbf{Comparative retrieval performance of \textit{CLEAR-WSI} across CAMELYON16, BRACS-1, BRACS-2, and X-HER2, together with available results from existing WSI retrieval methods.}
    Each task sector reports normalized discounted cumulative gain (NDCG@$k$) and precision at rank $k$ (P@$k$, i.e., Precision@$k$) for $k \in \{1,3,5\}$. Self-Retrieval (SR) is the fully automated \textit{CLEAR-WSI} configuration and does not require the query ground-truth label. Class-Informed (CI) uses the query ground-truth label and is therefore shown as a dashed oracle reference. For visualization, CAMELYON16, BRACS-1, and X-HER2 use a radial maximum of 100, whereas BRACS-2 uses a radial maximum of 80; values should therefore be compared using the labeled scale within each task sector.}
    \label{fig:main_results_radar}
\end{figure*}

\noindent\textbf{Comparison with Existing Retrieval Methods.}
Figure~\ref{fig:main_results_radar} compares \textit{CLEAR-WSI} with Yottixel-K, SISH, and RetCCL across the four retrieval tasks. In the label-unavailable setting, CLEAR-WSI (SR) achieves the highest $\mathrm{Precision}@5$ on CAMELYON16, improving over the strongest baseline by 3.10 \% (89.92\% vs. 86.82\%), while remaining competitive in $\mathrm{NDCG}@5$. It also consistently outperforms all baselines on both BRACS tasks across the reported neighborhood sizes. At $k=5$, the Precision gains over RetCCL are 2.75 \% on BRACS-1 (71.03\% vs. 68.28\%) and 1.95 \% on BRACS-2 (46.78\% vs. 44.83\%). Together with the consistent NDCG improvements, these results indicate that diagnostic-driven aggregation and SR filtering remain effective as class granularity increases.

On X-HER2, CLEAR-WSI (SR) matches the strongest baselines at $k=1$, is slightly lower at $k=3$, and surpasses them at $k=5$, where its $\mathrm{NDCG}@5$ improves over RetCCL by 6.18 \% (77.93\% vs. 71.75\%). This pattern suggests that SR provides a clearer advantage when a broader set of clinically relevant neighbors is retrieved. CI remains the strongest configuration across all neighborhood sizes, showing the potential benefit of an available diagnostic label. Detailed numerical results are reported in Table~\ref{tab:comparative_results}.

\noindent\textbf{Retrieval Stability Across Neighborhood Sizes ($k$).}
The SR configuration remains robust across $k \in \{1,3,5\}$. At $k=1$, it attains 90.02\% for both Precision and NDCG on CAMELYON16, exceeding both SISH and RetCCL. It also consistently leads all evaluated baselines on both BRACS tasks and at every reported $k$. On CAMELYON16, the only exception is $\mathrm{NDCG}@5$, where SR remains competitive with SISH. Overall, the results show that \textit{CLEAR-WSI} retrieves diagnostically relevant slides early in the ranked list, with its largest gains observed on the more heterogeneous BRACS tasks.

\noindent\textbf{Filtering Dynamics: CI vs. SR.}
CI filtering uses the query ground-truth label and is therefore reported as an oracle, label-available setting rather than a fully automated configuration. It consistently exceeds SR across all four tasks. At $k=5$, CI reaches 98.93\% Precision and 97.23\% NDCG on CAMELYON16; 80.69\% and 81.59\% on BRACS-1; 58.24\% and 65.20\% on BRACS-2; and 97.50\% and 96.54\% on X-HER2, respectively. The gap between CI and SR quantifies the potential benefit of a more accurate diagnostic prior. In contrast, SR requires no query ground-truth label and remains the fully automated configuration.

%% file: components/ablation_study.tex
To better understand CLEAR-WSI, we analyze its architecture, retrieval mechanisms, and computational efficiency. We first compare feature encoders and slide-level aggregation strategies, then evaluate the filter strategies and growing memory bank, and finally report latency at the tested memory-bank sizes.

\subsection{Architecture Ablation Study}
Table \ref{tab:ablation_architecture} compares feature encoders and WSI aggregation strategies. The results indicate that pathology-specific representations and diagnosis-guided slide aggregation are complementary for whole-slide retrieval.

\noindent\textbf{Impact of Feature Encoders.} We compare DeiT, MoCo v3, the Prov-GigaPath tile encoder, and UNI. While MoCo v3 and Prov-GigaPath provide incremental improvements over DeiT, UNI increases average NDCG by 20.51 percentage points relative to Prov-GigaPath. Within this CAMELYON16 ablation, UNI therefore provides the most effective patch representation for diagnosis-aligned retrieval.

\noindent\textbf{Selection of WSI Aggregation.}
Replacing linear principal component analysis (PCA) aggregation with DX-Agg,
an AttentionMIL module, consistently improves both retrieval metrics.
For example, DX-Agg with the DeiT encoder improves average NDCG from
63.66\% to 66.86\%.
The further gain obtained by combining UNI with DX-Agg shows that
pathology-specific patch features and diagnosis-guided slide aggregation
are complementary.

\newsavebox{\tablebox}

\begin{table*}[t] \centering \caption{Architecture ablation on CAMELYON16. The table evaluates the impact of feature encoders and WSI aggregation methods on retrieval performance. Results represent the average NDCG and Precision across $k \in \{1,3,5\}$.} \label{tab:ablation_architecture} \begin{lrbox}{\tablebox} \begin{tabular}{l c c c c c c c c} \toprule \fontsize{8}{10}\selectfont & \multicolumn{4}{c}{Feature Encoder} & \multicolumn{2}{c}{WSI Aggregator} & \multicolumn{2}{c}{CAMELYON16} \\ \cmidrule(lr){2-5} \cmidrule(lr){6-7} \cmidrule(lr){8-9} Configuration & DeiT & MoCo v3 & Prov-GigaPath & UNI & PCA & DX-Agg & \makecell[c]{Average\\NDCG} & \makecell[c]{Average\\Precision} \\ \midrule DeiT + PCA & \checkmark & & & & \checkmark & & 63.66 & 66.41 \\ DeiT + DX-Agg & \checkmark & & & & & \checkmark & 66.86 (\greenup{3.20}) & 69.00 (\greenup{2.59}) \\ MoCo v3 + DX-Agg & & \checkmark & & & & \checkmark & 67.06 (\greenup{0.20}) & 70.54 (\greenup{1.54}) \\ Prov-GigaPath + DX-Agg & & & \checkmark & & & \checkmark & 69.04 (\greenup{1.98}) & 70.03 (\reddown{0.51}) \\ \textbf{UNI + DX-Agg (CLEAR-WSI)} & & & & \checkmark & & \textbf{\checkmark} & \textbf{89.55} (\greenup{20.51}) & \textbf{89.66} (\greenup{19.63}) \\ \bottomrule \end{tabular} \end{lrbox} \scalebox{0.9}{\usebox{\tablebox}} \end{table*}

\subsection{Retrieval Mechanism Ablation Study} 
Table \ref{tab:ablation_mechanism} evaluates the filter strategy and growing memory bank used by CLEAR-WSI. The fixed bank is included only as an ablation comparator to isolate the contribution of incremental case accumulation; the proposed system uses the growing configuration. We conduct this analysis on BRACS Level-2 because retrieval performance on CAMELYON16 and X-HER2 is close to saturation, making the effects of individual retrieval components less distinguishable.

\noindent\textbf{Impact of Filter Strategy.} Applying an SR filter (prediction-based) consistently improves retrieval quality over the unfiltered baseline under the label-unavailable setting. Restricting retrieval to diagnostically relevant candidates reduces semantic mismatch and improves neighborhood consistency. When labels are available, retrieval performance improves substantially further, indicating that retrieval quality is strongly influenced by the quality of the class prior used for candidate selection. The gap between prediction-based and ground-truth filtering (CI filter) therefore provides an estimate of the potential room for improvement achievable through more accurate diagnostic priors. 

\noindent\textbf{Contribution of the Growing Memory Bank.} Introducing the growing memory bank further improves retrieval performance under both label-available and label-unavailable settings. These gains suggest that continuously accumulating previously observed cases provides a richer and more diverse retrieval space, allowing the system to identify more diagnostically relevant neighbors over time. Notably, the improvement is observed regardless of the filtering strategy, indicating that the benefit of the growing memory bank is complementary to filter quality rather than dependent on it.

\noindent\textbf{Combined Effect.} The strongest results are obtained by combining CI filtering with the growing memory bank in the oracle setting. The same benefit from memory-bank growth is observed with SR when query labels are unavailable, showing that incremental case accumulation complements both filtering strategies within the evaluated setting.

\begin{table*}[t] \centering \caption{Mechanism ablation on BRACS. The table evaluates the contribution of filter strategies and the proposed growing memory bank under both automated and oracle settings. NDCG and Precision (\%) are averaged over $k \in \{1,3,5\}$. CI Filter: assuming ground-truth labels are available. } \label{tab:ablation_mechanism} \begin{lrbox}{\tablebox} \begin{tabular}{l l c c c c c c c} \toprule \fontsize{8}{10}\selectfont & & \multicolumn{3}{c}{Filter Strategy} & \multicolumn{2}{c}{Memory Bank} & \multicolumn{2}{c}{BRACS Level-2} \\ \cmidrule(lr){3-5} \cmidrule(lr){6-7} \cmidrule(lr){8-9} Setup & Pipeline & \makecell[c]{No\\Filter} & SR & CI & Fixed & Growing & \makecell[c]{Average\\NDCG} & \makecell[c]{Average\\Precision} \\ \midrule \multirow{3}{*}{\makecell[c]{Label\\Unavailable}} & UNI + DX-Agg & $\checkmark$ & & & $\checkmark$ & & 41.72 & 41.40 \\ & UNI + DX-Agg + SR Filter & & $\checkmark$ & & $\checkmark$ & & 44.82 (\greenup{3.10}) & 43.99 (\greenup{2.59}) \\ & \textbf{UNI + DX-Agg + SR Filter + Growing Bank} & & $\checkmark$ & & & $\checkmark$ & \textbf{46.55} (\greenup{1.73}) & \textbf{46.37} (\greenup{2.38}) \\ \midrule \multirow{2}{*}{\makecell[c]{Label\\Available}} & UNI + DX-Agg + CI Filter & & & $\checkmark$ & $\checkmark$ & & 55.31 (\greenup{13.59}) & 52.92 (\greenup{11.52}) \\ & \textbf{UNI + DX-Agg + CI Filter + Growing Bank} & & & $\checkmark$ & & $\checkmark$ & \textbf{60.97} (\greenup{5.66}) & \textbf{57.79} (\greenup{4.87}) \\ \bottomrule \end{tabular} \end{lrbox} \scalebox{0.75}{\usebox{\tablebox}} \end{table*}

\subsection{Latency and Scalability Analysis}
Table~\ref{tab:latency} reports inference efficiency on an NVIDIA H100 graphics processing unit (GPU). Feature extraction incurs a one-time offline cost of 2646.99 seconds for CAMELYON16, 3209.60 seconds for BRACS, and approximately 472.01 seconds for X-HER2. During sequential evaluation, the memory bank grows from 27 to 156 entries on CAMELYON16, from 64 to 151 entries on BRACS, and from 18 to 34 entries on X-HER2; feature loading and retrieval require 7.17, 5.83, and 1.72 seconds, respectively, for the complete test sets. These measurements demonstrate lightweight online retrieval at the evaluated memory-bank sizes, while larger hospital-scale archives require further validation.


\begin{table*}[t] \centering \caption{Latency and scalability analysis of \textit{CLEAR-WSI} on an NVIDIA H100 GPU. The memory bank grows after each test slide is processed. Results are reported with $K=5$, and online latency is computed as feature loading plus retrieval.} \label{tab:latency} \begin{lrbox}{\tablebox} \begin{tabular}{l c c c c c c} \toprule \fontsize{8}{10}\selectfont & \multicolumn{2}{c}{Dataset Profile} & \multicolumn{1}{c}{Offline Stage} & \multicolumn{3}{c}{Online Stage} \\ \cmidrule(lr){2-3} \cmidrule(lr){4-4} \cmidrule(lr){5-7} Dataset & Slides & \makecell[c]{Memory Bank\\Size} & \makecell[c]{Feature\\Extraction (s)} & \makecell[c]{Feature\\Loading (s)} & Retrieval (s) & \makecell[c]{Online\\Total (s)} \\ \midrule CAMELYON16 & 399 & $27 \rightarrow 156$ & \makecell[c]{2646.99\\(6.6/slide)} & 3.55 & 3.62 & \textbf{7.17} \\ BRACS & 544 & $64 \rightarrow 151$ & \makecell[c]{3209.60\\(5.9/slide)} & 3.17 & 2.66 & \textbf{5.83} \\ X-HER2 & 77 & $18 \rightarrow 34$ & \makecell[c]{$\sim$472.01\\(6.13/slide)} & 1.49 & 0.23 & \textbf{1.72} \\ \bottomrule \end{tabular} \end{lrbox} \scalebox{0.85}{\usebox{\tablebox}} \end{table*}

%% file: components/discussion.tex
\noindent\textbf{Performance Across Retrieval Tasks.}
The retrieval results indicate that the advantage of \textit{CLEAR-WSI} becomes more apparent as the diagnostic difficulty of the task increases. On CAMELYON16, several methods already operate near a high-performance regime, and \textit{CLEAR-WSI} (SR) is competitive with SISH for NDCG@$5$ while providing higher Precision@$5$. The separation becomes clearer on BRACS, where subtyping produces more heterogeneous and fine-grained neighborhoods. SR improves both ranking quality and precision over the evaluated baselines across the two BRACS tasks. This suggests that diagnosis-driven slide aggregation is especially useful when diagnostically relevant evidence is distributed across multiple tissue regions.

\noindent\textbf{Contribution of Each System Component.}
The ablation study shows that retrieval performance benefits from three complementary components: slide representation, retrieval refinement, and archive growth. Replacing PCA with DX-Agg consistently improves retrieval quality when the underlying patch features are held fixed, indicating that diagnosis-driven aggregation contributes independently of the feature encoder. SR further improves retrieval by enforcing diagnostic consistency among retrieved candidates, while the growing memory bank provides additional gains for both SR and CI settings. Together, these results suggest that the performance improvements arise from the combined effect of stronger slide representations and retrieval-stage refinement rather than any single component alone.

\noindent\textbf{Latency and Scalability.}
The latency analysis shows that most computation is incurred during the one-time feature extraction stage, whereas retrieval itself remains efficient at the evaluated memory-bank sizes. Since each indexed WSI contributes a single embedding and diagnostic label, the storage required for archive growth remains constant per slide.
This compact representation reduces storage overhead and simplifies archive expansion as new cases are added. Integrated with the growing memory-bank design and efficient retrieval, it provides a practical foundation for deployment in continuously expanding clinical archives.

\noindent\textbf{Limitations and Future Work.}
The present study is limited to three breast-pathology datasets, four diagnostic tasks, and memory banks containing hundreds rather than millions of WSIs. Although CLEAR-WSI achieves efficient retrieval with a constant per-slide storage footprint, the reported latency measurements were obtained on a single hardware configuration and may not generalize to large-scale clinical infrastructures. In addition, retrieval relevance was defined by dataset labels and therefore may not fully reflect clinically meaningful morphological similarity. Future work will focus on large-scale multi-organ validation, robustness to distribution shifts, and reader studies evaluating the impact of retrieval on diagnostic practice.

%% file: components/appendix.tex
Table~\ref{tab:comparative_results} reports the complete NDCG@$k$ and Precision@$k$ values underlying the radar visualization in Fig.~\ref{fig:main_results_radar} for all evaluated methods, tasks, and neighborhood sizes $k \in \{1,3,5\}$.

\begin{table*}[t]
\centering
\caption{Complete comparative benchmarking of \textit{CLEAR-WSI} and existing retrieval methods across four retrieval tasks. Results are reported as NDCG@$k$ and Precision@$k$. CI uses the query ground-truth label and is reported as an oracle reference. The best result is shown in \textbf{bold}, and the second-best result is \underline{underlined}.}
\label{tab:comparative_results}
\begin{adjustbox}{max width=\textwidth}
\begin{tabular}{llcccc}
\toprule
& Methods & CAMELYON16 & BRACS-1 & BRACS-2 & X-HER2 \\
& & NDCG@$k$ $\vert$ Precision@$k$ & NDCG@$k$ $\vert$ Precision@$k$ & NDCG@$k$ $\vert$ Precision@$k$ & NDCG@$k$ $\vert$ Precision@$k$ \\
\midrule
\multirow{5}{*}{\rotatebox{90}{$k=1$}}
& Yottixel-K \cite{kalra2020yottixel} & 76.21$\vert$76.21 & 49.91$\vert$49.91 & 33.61$\vert$33.61 & 68.75$\vert$68.75 \\
& SISH \cite{chen2022fast} & 86.66$\vert$86.66 & 57.47$\vert$57.47 & 37.93$\vert$37.93 & \underline{81.25$\vert$81.25} \\
& RetCCL \cite{wang2023retccl} & 85.32$\vert$85.32 & 63.22$\vert$63.22 & 42.53$\vert$42.53 & \underline{81.25$\vert$81.25} \\
& CLEAR-WSI (SR UNI) & \underline{90.02$\vert$90.02} & \underline{67.82$\vert$67.82} & \underline{44.83$\vert$44.83} & \underline{81.25$\vert$81.25} \\
& CLEAR-WSI (CI UNI) & \textbf{96.93$\vert$96.93} & \textbf{86.21$\vert$86.21} & \textbf{56.32$\vert$56.32} & \textbf{93.75$\vert$93.75} \\
\midrule
\multirow{5}{*}{\rotatebox{90}{$k=3$}}
& Yottixel-K \cite{kalra2020yottixel} & 74.59$\vert$78.29 & 49.58$\vert$54.12 & 31.25$\vert$36.47 & 64.25$\vert$64.58 \\
& SISH \cite{chen2022fast} & 86.90$\vert$86.25 & 60.34$\vert$59.00 & 44.56$\vert$43.68 & 68.28$\vert$68.75 \\
& RetCCL \cite{wang2023retccl} & 86.47$\vert$86.18 & 68.03$\vert$68.58 & 43.26$\vert$44.44 & \underline{80.75$\vert$81.25} \\
& CLEAR-WSI (SR UNI) & \underline{89.56$\vert$89.15} & \underline{71.26$\vert$71.65} & \underline{47.14$\vert$47.51} & 77.93$\vert$77.08 \\
& CLEAR-WSI (CI UNI) & \textbf{97.69$\vert$99.22} & \textbf{81.46$\vert$80.46} & \textbf{61.39$\vert$58.81} & \textbf{95.22$\vert$95.83} \\
\midrule
\multirow{5}{*}{\rotatebox{90}{$k=5$}}
& Yottixel-K \cite{kalra2020yottixel} & 73.69$\vert$77.49 & 48.64$\vert$54.12 & 29.17$\vert$36.47 & 64.17$\vert$63.75 \\
& SISH \cite{chen2022fast} & 89.67$\vert$85.01 & 67.67$\vert$66.90 & 43.46$\vert$43.21 & 69.08$\vert$68.75 \\
& RetCCL \cite{wang2023retccl} & 86.05$\vert$86.82 & 68.39$\vert$68.28 & 44.07$\vert$44.83 & 71.75$\vert$72.50 \\
& CLEAR-WSI (SR UNI) & \underline{89.07$\vert$89.92} & \underline{71.56$\vert$71.03} & \underline{47.67$\vert$46.78} & \underline{77.93$\vert$76.25} \\
& CLEAR-WSI (CI UNI) & \textbf{97.23$\vert$98.93} & \textbf{81.59$\vert$80.69} & \textbf{65.20$\vert$58.24} & \textbf{96.54$\vert$97.50} \\
\bottomrule
\end{tabular}
\end{adjustbox}
\end{table*}